\pdfoutput=1

\documentclass[11pt]{article}

\usepackage[review]{EMNLP2023}
\usepackage{times}
\usepackage{latexsym}
\usepackage{hyperref}
\usepackage[T1]{fontenc}
\usepackage{amssymb}
\usepackage{verbatim}
\usepackage{xcolor}
\usepackage{listings}
\usepackage{subfigure}
\usepackage{makecell}
\usepackage{mathrsfs}
\usepackage{utfsym}
\usepackage[normalem]{ulem}
\useunder{\uline}{\ul}{}
\usepackage[inline]{enumitem}
\usepackage{enumitem}
\usepackage[utf8]{inputenc}

\usepackage{microtype}

\usepackage{inconsolata}
\usepackage{graphicx}

\title{AgentSims: An Open-Source Sandbox for Large Language Model Evaluation }

\author{Jiaju Lin$^{1,2}$, Haoran Zhao$^{1,3}$ \thanks{$^*$ Corresponding author.} , Aochi Zhang$^{1}$,  Yiting Wu$^{1,4}$, \\
\textbf{Huqiuyue Ping$^{1,5}$, Qin Chen$^{6}$} \\
$^1$PTA Studio \\
$^2$ Pennsylvania State University,
$^3$ Beihang University, \\
$^4$ Sun Yat-sen University,
$^5$Zhejiang University,
$^6$East China Normal University \\
$^3$zhaohaoran@buaa.edu.cn \\$^2$jjlin.unfake@gmail.com  \and $^6$qchen@cs.ecnu.edu.cn\\
}

\begin{document}
\nolinenumbers
\maketitle

\begin{abstract}

With ChatGPT-like large language models (LLM) prevailing in the community, how to evaluate the ability of LLMs is an open question. Existing evaluation methods suffer from following shortcomings: (1) constrained evaluation abilities, (2) vulnerable benchmarks, (3) unobjective metrics. We suggest that task-based evaluation, where LLM agents complete tasks in a simulated environment, is a one-for-all solution to solve above problems. We present AgentSims, an easy-to-use infrastructure for researchers from all disciplines to test the specific capacities they are interested in.
Researchers can build their evaluation tasks by adding agents and buildings on an interactive GUI or deploy and test new support mechanisms, i.e. memory, planning and tool-use systems, by a few lines of codes. Our demo is available at \url{https://agentsims.com} . 

\end{abstract}

\section{Introduction}\label{intro}

LLMs have revolutionized Natural Language Processing (NLP) and beyond. They demonstrate great potential in few-shot learning\cite{brown2020GPT3}, code generation\cite{nijkamp2023codegen}, reasoning\cite{yao2023ToT} and other tasks. Furthermore, LLM powered autonomous agents\cite{weng2023AutoAgents} are widely applied in solving complex problems, like multimodal generation\cite{shen2023hugginggpt}, software developing\cite{qian2023chatdev} and social simulating \cite{park2023generative}.

Although LLMs have reformed the paradigm of NLP, the problem of evaluation keeps haunting this field. Old benchmarks become out-of-date. Since LLMs achieve human-level Natural Language Understanding (NLU) and Natural Language Generation (NLG) abilities\cite{openai2023gpt4}. 
To address the pressing need for novel benchmarks, the NLP community has introduced an array of fresh evaluation tasks and datasets, encompassing a diverse spectrum of abilities, including close-book question-answering (QA) based knowledge testing\cite{ hendrycks2020MMLU, huang2023ceval}, human-centric standardized exams\cite{zhong2023agieval}, multi-turn dialogue\cite{lin2023llmeval}, reasoning\cite{liu2023reasonAbility,srivastava2023big-bench} and safety assessment\cite{sun2023safetyAssess}.

However, there are still many problems with these new benchmarks. 
1) Evaluated abilities are limited by the task formats.
Since a majority of these tasks adopt a single-turn QA format, they are insufficient to comprehensively evaluate various aspects of LLMs' capabilities. For instance, they fail to assess the models' proficiency in adhering to instructions in dialogue or mimicking human-like social interactions.
2) Benchmarks can be easily hacked. Avoiding the leakage of test set is of paramount importance when evaluate a model's ability. Nonetheless, considering the amount of pretrained knowledge of LLM, it has become more and more inevitable to inadvertently mix test cases into the training set.\cite{gunasekar2023textbooks}. 
3) For open-ended QA, existing metrics are not objective. 
Previous metrics for open-ended QA involve automatic metrics, and human-rating as subjective metrics\cite{zhou2023lima}. 
In the LLM era, text segment matching based metrics become out-of-date. To mitigate the high-costly issue of human-rating, today's researchers employ well-aligned LLMs like GPT4 as automatic raters. Nevertheless, the most significant problem of this approach is that it can not evaluate super GPT4-level models, and LLMs are biased toward specific features \cite{wang2023chatgptNLGeval}.

Based on these observations, we suggest task-based evaluation for LLM benchmarks. Specifically, given an artificial social-economic environment, LLM-driven agents should achieve the pre-defined task goals to prove their abilities, just like humans accomplishing goals in real world or games to show their capacities.
Task-based evaluation is a one-for-all solution for current issues: 
1) Task-based evaluation can test an LLM's overall ability. The complexity of social simulation and adaptation far exceeds simple QA and can formulate more challenging tasks for LLMs.  LLM agents need to be equipped with the ability from NLU to Theory of Mind (ToM) \cite{premack1978ToM}.  
2) Task solving processes are less likely to be hacked. Different from unchanged test datasets whose formats can be easily mimicked and added to training data. Task settings are diversified and the emergent social behaviors and groups are less likely to be described and included in training corpus. 
3) Task passing rate is an objective metric. 
Compared with popular rating methods by ChatGPT, the passing rate does not rely on any black-box rating process, i.e. deep neural networks or human brains, thus it is an objective and fair metric for the comparison between LLMs.



To all-around estimate LLMs' capacities, we hope researchers from all fields take part in the development of evaluation tasks. 
However, a key obstacle to fostering a collaborative research community is the absence of a standard paradigm, an easy-to-use and extensible research platform. Previous works pursue the most efficient way to implement a sandbox while ignoring the need of non-specialist users. Besides, the poor readability further results in poor extensiblity and user churn. Moreover, the agents' performance varies with different support systems, i.e. memory, planning and tool-use system. We need a standard implementation to ensure the reproducibility of experimental results.

To this end, we introduce AgentSims, an interactive, visualized, and program-based infrastructure for curating evaluation tasks for LLMs. It creates an artificial town with various buildings and residents. The core objective of AgentSims is to streamline the task design process, eliminating hurdles that researchers from various backgrounds and programming proficiencies might encounter.
\begin{itemize}[leftmargin=*, align=left]
\vspace{-1mm}
    \item For researchers focusing on LLM, AgentSims is \textbf{extendable and combinable} to allow users to combine different plan, memory and learning systems to study the impacts and effectiveness of various system design. 
    \vspace{-1mm}
    \item For experts from other fields like behavioral economics or social psychology, AgentSims provides \textbf{an interactive UI} for map design and agent creation and lower the entry threshold. Such a user-friendly architecture further facilitates the cooperation between different fields and the future prosperity of the LLM community.  
\end{itemize}



\section{Related Work}
\subsection{Benchmarks for Large Language Models}
The emergency of ChatGPT and other LLMs requires new benchmarks for effective evaluation. \citet{srivastava2023big-bench} is the most accepted benchmark to evaluate LLM's general abilities. It contains more than 200 tasks, covering from childhood development, to social bias. \citet{zhong2023agieval} collect test tasks from human-centric standardized exams like GRE and SAT. 
\cite{hendrycks2020MMLU,huang2023ceval} are benchmarks focusing on measuring knowledge acquired in pretraining. They covers subjects across STEM, the humanities, the social sciences.
\citet{lin2023llmeval} build a benchmark for LLMs' multiturn dialogue abilities. Every dialogue is limited to two turns for simplicity.
\citet{sun2023safetyAssess} focus on measure the safety of LLMs. They curate a adversarial attack dataset containing insulting instructions and test whether LLMs can be jailbroke. 
However, as mentioned above, existing datasets have issues that can not fully demonstrate abilities of LLMs. AgentSims overcomes these difficulties and renders a chance for overall evaluation of LLMs.



\begin{figure*}[!h]\centering
    \centering
    \includegraphics[width=\linewidth]{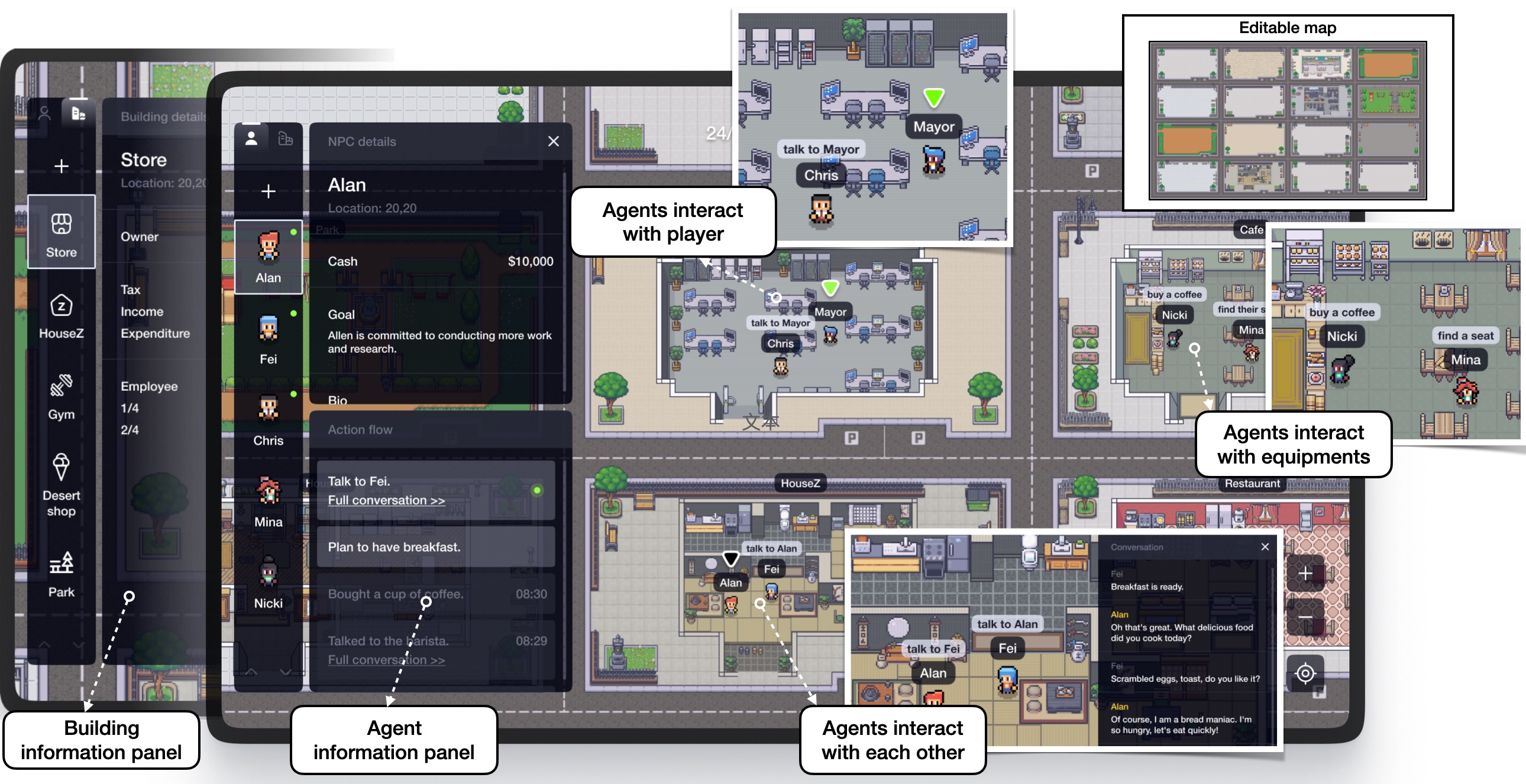}
    \caption{Front end of AgentSims, showing in a pixel game style. Users can create agents and buildings in the left-side panel and observe agents behaviors in the main screen. Besides setting-then-observing, users can also play as the mayor and talk with agents to intervene the experiment.}
    \label{fig:front-end}
\end{figure*}

\subsection{Multi Agent Cooperation}
With LLMs demonstrate their overwhelming abilities, researchers find that multi LLM agents can generate better results than a single one. 
\citet{nair2023dera} is one of the earliest attempts of multi-agent cooperation. It builds a forum for agents to communicate feedback and iteratively improve their healthcare suggestions. \citet{li2023camel} expand the application field of agent cooperation method by role-playing. From programming to domain-specific QA, it surpass single agent baselines. \citet{qian2023chatdev} build a software development company, by meticulously dividing the development process into four distinct stages, leading to efficient resolution of specific subtasks. \citet{liu2023stableAlign} first apply multi-agent simulated society for alignment, where agents in a sandbox learn from social interaction to understand moral rules.
\cite{park2023generative} is the most sophisticated application of multi agent sandbox. Authors build support mechanisms to enable agents to produce believable individual and emergent social behaviors. However, none existing methods provide a user-friendly interface for unprofessional researchers or build a standard paradigm for agent support system. 
Nonetheless, current multi-agent systems are task-oriented rather than evaluation-oriented. AgentSims works as a platform for easy benchmark construction.

\section{Key Components}

\begin{figure*}[!h]\centering
    \centering
    \includegraphics[width=\linewidth]{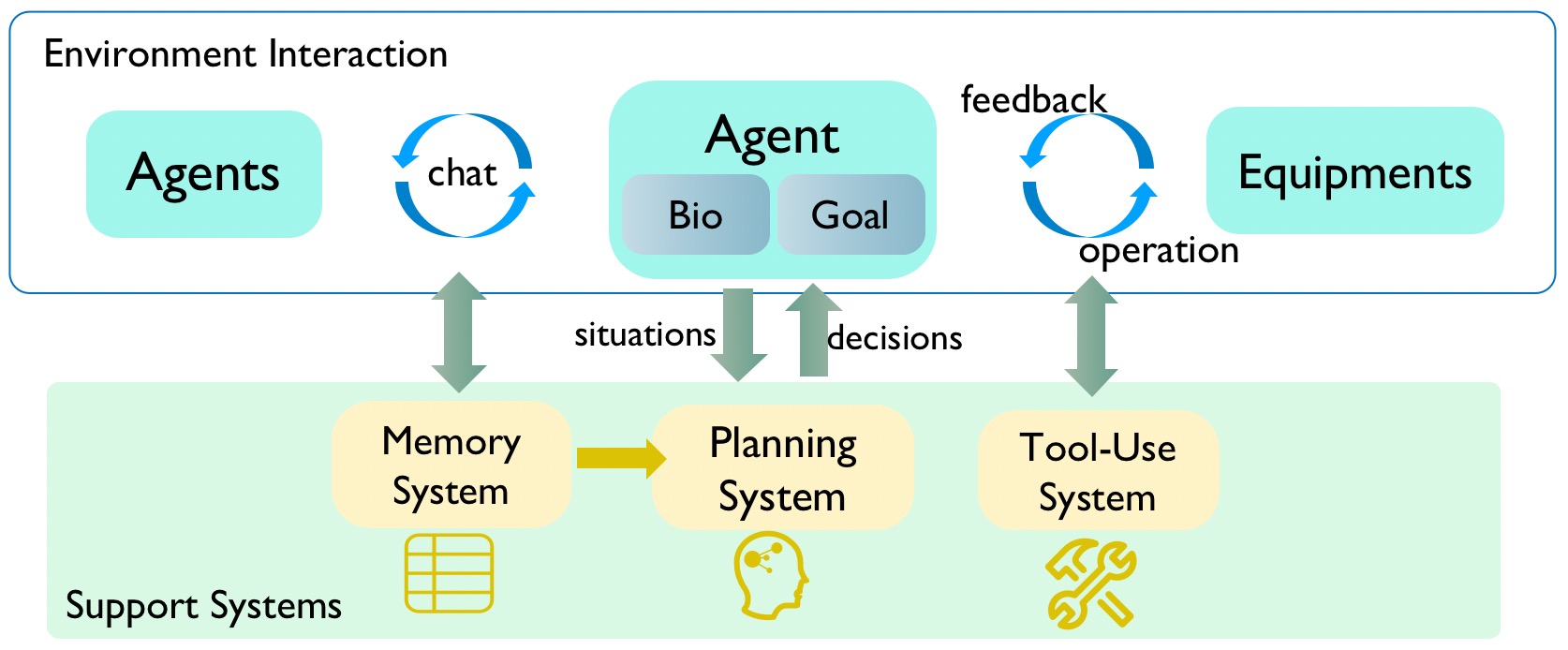}
    \caption{Overview of AgentSims architecture}
    \label{fig:prompt-achi}
\end{figure*}

As shown in Figure~\ref{fig:prompt-achi}, key components of AgentSims can be divided into two parts: 1) generative agents driven by LLM support mechanisms. 2) buidlings and equipment that consist the sandbox environment. 

\subsection{Generative Agents}

If prompted properly, LLMs can generate believable behaviors\cite{park2022socialSim}. However, to achieve human-like memory performance and long-term coherence, LLM is not enough. We need auxiliary systems to enable agents to perform more naturally. Referring to recent work\cite{park2023generative, wang2023voyager}, we abstract these supportive mechanisms into three parts: Planning System, Memory System, and Tool-Use System.

\textbf{Planning System} 
LLMs have shown some planning and reasoning capacities. However, faced with complex tasks, vanilla LLMs always fail for lacking long-term arrangement abilities. Hence, we introduce a Planning System to ensure agents' behaviors are coherent and believable. The Planning System reorganizes a goal by decomposing the target, summarizing current condition and generating subtasks. Specifically, it is assembled by a series of pluggable prompt modules, which assess current achievement of ultimate goals by checking the memory system and making decisions for next steps. Once a new step is completed, it would be recorded in the memory system.

\textbf{Memory System.} Agents capable of emulating human behavior necessitate comprehending a vast array of experiences, beyond what a prompt can contain. The complete memory stream is too expensive to be accommodated in the limited context window, and attempting to do so can overwhelm the model. Thus, we add a memory system for agents' experience retention and retrieval. The system is built upon a vector database for efficient storing and retrieving. Specifically, every agent's daily memory is encoded into embeddings and stored in the database. Every time when agents face some new situation that needs the previous memory, such as chatting with familiar people, the memory system can retrieve the information about their relationship to improve agent behaviour consistency. 


\textbf{Tool-Use System.} Ideally, agents continuously explore the simulated world would learn from previous failures and successes, then acquire diverse skills. In our framework, to realize this feature, we present a tool-use system, which endows agents with the ability to accomplish real-world tasks. Particularly, the tool use system stores equipment-operation pairs learning from feedback of using equipment. Once agents select equipment to interact with by planning and memory system, they need to infer an initial operation by the description of the equipment. And the equipment will return an operation result as feeedback. If the agent believes the result meets their operation purpose, a new skill would be stored in the Tool-Use System. 


\subsection{Buildings and Equipment}
Interactive buildings and equipment are necessities for the diversity of an LLM sandbox. They compose the physical environments of the simulated world. In our framework, a building or location contains equipment like stoves or office desks. Thus, buildings are defined by the equipment they contain and equipment is the basic element composing the interactive environment.
More specifically, the equipment can be defined by some definition texts describing its features and support function,
which can be either hard-coded by the developer or a language model that supports self-adaptive agent-equipment interaction.
When an agent interacts with equipment, as shown in Figure~\ref{fig:prompt-achi}, its operation text will be sent to the background support model. The support function then returns the operation outcome based on the predefined rules or model-generated texts. For example, if an agent wants to get a cup of tea from a stove, the operation is 'Get a cup of tea' and the support function may return 'Meaningless operation' according to the hard code or 'You can not get tea from a stove' generated by the model. Then the agent would learn from the feedback and refine its operations. 

\section{Interaction scenarios}
Regarding the researchers' backgrounds and purposes, we design two interaction modes: User Mode and Developer Mode. In the User Mode, researchers who consider little about background support systems are target users. For researchers chasing better LLMs performance, Developer Mode provides flexible protocols for their development of different support mechanisms.
\subsection{User Mode}
In the User Mode, AgentSims provides an interactive interface in a pixel game style, as shown in Figure \ref{fig:front-end}. Researchers can create agents, construct buildings and equipment in a graphical interface, focusing on the rationality of experiment design, free from complex background driving mechanisms.

\begin{figure}[!t]\centering
    \centering
    \includegraphics[width=\linewidth]{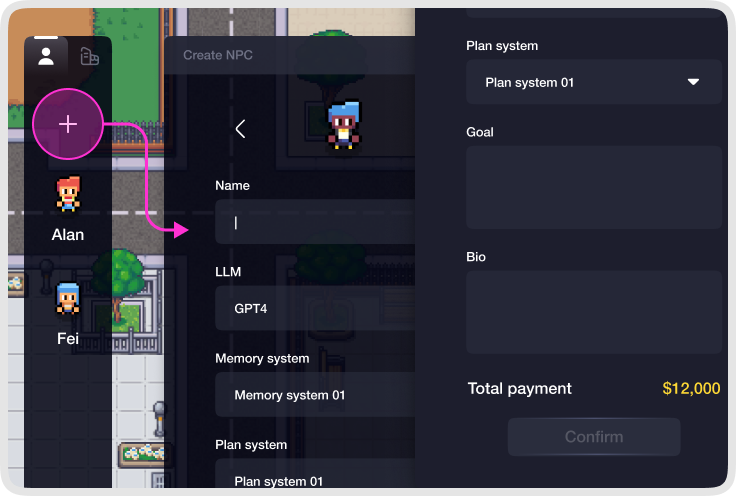}
    \caption{Agent Creation}
    \label{fig:NPC creation}
\end{figure}

\textbf{Agent Creation.}
Users can define agents within the system through an easy-to-use front end, as shown in the Figure \ref{fig:NPC creation}. AgentSims provides various protocols for users to create functional agents. Not only basic information like goals and biography, but also options of Memory and Planning Systems. We pre-design a list of memory and planning systems and users can choose their preference from a drop-down menu. 

\begin{figure}[!t]\centering
    \centering
    \includegraphics[width=\linewidth]{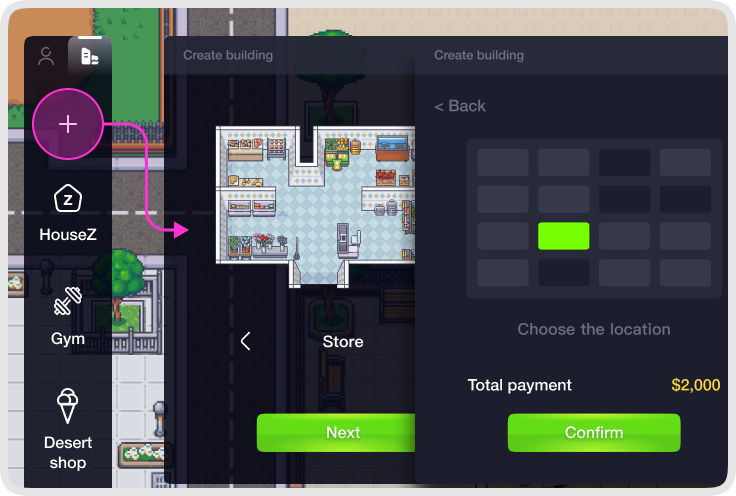}
    \caption{Building Creation}
    \label{fig:Builidng creation}
\end{figure}

\textbf{Building Creation.}
Users can also customize the physical environment by constructing  buildings. As shown in Figure~\ref{fig:Builidng creation}, users define a building by choosing a pre-configured building with equipment inside. To be noticed, the equipment in buildings are predefined but can be modified in the Developer Mode.

\textbf{Experiment Intervene.}
Besides observing, users can play as the major agent to participate in the experiment. By talking with other agents, users can intervene the experiment naturally rather than modify agents' memory or goals roughly. 

\subsection{Developer Mode}
Developer Mode is designed for professional developers who are familiar with the properties of LLMs and pursue better performance of LLMs on a well-defined complex task. The highly-modularized feature of AgentSims enables developers to add new functions within a few lines of code.

\textbf{Agent Design.}
Developers have the flexibility to create agents tailored for various objectives and assemble diverse agents within a single sandbox for observation. To streamline the process of agent customization, we've abstracted the LLM backbone and distinct support systems into separate classes and function calls, as illustrated below. This empowers developers to personalize an agent by making adjustments to these abstract functions.
\linebreak
\linebreak

\begin{lstlisting}
class LLMCaller:
    def __init__(self, model: str) -> None:
        self.model = get_model(model)
    
    def ask(self, prompt: str) :
        result = self.model.generate(prompt)
        return result

class Agent:
    def __init__(self, name, bio, goal, model, memorySystem, planSystem, buildings, cash):
        self.state = State()
        self.state.buildings = buildings
        self.state.cash = cash
        self.caller = Caller(model)
        
    def plan(self) -> None:
        self.state.plan_prompt = ...
        self.state.plan = self.caller.ask(self.state.pl_prompt)

    def memory_store(self) -> None:
        self.state.memory_prompt = ...
        self.state.memory = 
        self.caller.ask(self.state.mem_prompt)

    def use(self, facility: str, operation: str, description: str) -> None:
        self.state.use_prompt = ...
        self.state.use = self.caller.ask(self.state.use_prompt)
\end{lstlisting}
\textbf{Building and Equipment Design.}
To customize the physical environment,  developers can design new buildings and equipment by configuring corresponding json files.\\
A new equipment can be defined by its type, description and a support function.
\begin{lstlisting}
[{"id": 1,
    "type": "counter",
    "function":..., 
    "description": "This is the counter ...",}]
\end{lstlisting}
In some cases, agents can purchase commodities or earn salaries at the equipment. We use another configure file to annotate these economic features.
\begin{lstlisting}
[{ "id": 1,
    "menu": {
        "chicken": 20,},
    "salary":0,}],
\end{lstlisting}
We define buildings by a type and the equipment it contains. Hence we use a two-dimensional array to mark the facility ids in the building blocks.
\begin{lstlisting}
[{"assets": "store_v1.2_0719",
    "id": 1,
    "price": 2000,
    "type": "store",
    "blocks":[[1,0,0...1,1]],
    "equipment":[0,1,0..]]}]
\end{lstlisting}



\section{Implementation}

AgentSims is run using Python 3.9\footnote{\url{https://www.python.org/downloads/release/python-390}} and requires installing the requirements.txt file provided in the codebase using Python's package manager PyPI\footnote{\url{https://pypi.org/}}.

\subsection{Backend}
The web server is built using Tornado\footnote{\url{https://www.tornadoweb.org/en/stable/}}, a lightweight Python web framework. It also uses the websockets library for API calls and push notifications, and mysql-connector-python to interact with the MySQL\footnote{\url{https://www.mysql.com/}} database. 

\subsection{Frontend}
Frontend The web client is built with Unity\footnote{\url{https://unity3d.com}}. The client built by WebGL\footnote{\url{https://get.webgl.org}} is embedded in the project code and can be accessed through a browser after proxying with nginx\footnote{\url{https://nginx.org/en/}}.

\section{Example Application Tasks}

\subsection{Subject LLM as participants}
When subject LLM agents are participants of an artificial scenario, researchers can evaluate LLM's social abilities, like ToM . In this case, the formulation of specific social scenes is realized by other baseline agents driven by stronger LLMs. For example, to study a new model's social adaptation abilities in a hostile environment, we can  embed colleague agents driven by GPT4 with a strong desire of bullying newcomers.  Then we place subject agents into this adversarial milieu and test whether the new model can understand other's emotion and improve how colleagues perceive it. 

%

\subsection{Subject LLM as mayor}
To assess LLM's long-term planning and organization abilities, researchers can appoint the subject LLM as the mayor of a town or the president of a company, where residents or employees are driven by baseline agents like GPT4. To overcome the difficulties set ahead deliberately or emerging during the experiments, then achieve the final goal of the task, the subject LLM needs to recruit new residents to handle new problems, issue sound policies and modify the out-of-date ones, found new functional buildings to satisfy emerging requirements, and so on. By analyzing the success rate of LLM mayor under different difficulties, researchers can gain valuable insights into the diverse capabilities of the LLM.

\subsection{Applications besides Evaluation}
Besides evaluating LLMs, AgentSims can be used as a data generation platform. Due to the fantastic NLG abilities of LLMs, researchers have applied them in data annotation and augmentation. However, some data involving social judgement and participation necessitate a more intricate approach than a single prompt can provide. Thus, we can simulate a specific social background and let LLMs generate data more precisely. \citet{liu2023stableAlign} have applied simulated society in alignment data generation. With AgentSims tailored for more intricate social simulations, its potential for enhancing data generation across various disciplines is undeniable.\\
Moreover, our program can also benefit social science researchers, by conducting more controllable preliminary experiments. Given that sota LLMs can understand human instructions and simulate human behaviours, social science researchers can design social environments as they wish for preliminary studies. Once researchers have a hypothesis, pilot experiments can be conducted in our virtual sandbox as a feasibility check.





\vspace{-1mm}
\section{Conclusion}
In this paper, we present AgentSims, avisualized and program-based infrastructure for LLM test sandbox construction.
AgentSims aims to facilitate researchers in effectively building LLM  evaluation tasks. It not only intends to make all its code openly available but also commits to continuously updating its documentation with comprehensive tutorials.
\vspace{-2mm}
\section*{Limitations}
\vspace{-1mm}
As a sandbox system, AgentSims' simulation ability is limited by the accuracy of LLMs and the diversity of buildings and equipment. It can never fully reflect real world cases. Besides, although task-based evaluation is a sound approach to measure the general ability of LLMs, it can hardly reflect fine-grained abilities like math reasoning. The pass rate of tasks can not provide insights on why LLMs success or fail. 

\bibliography{anthology,custom}

\begin{thebibliography}{24}
\expandafter\ifx\csname natexlab\endcsname\relax\def\natexlab#1{#1}\fi

\bibitem[{bench authors(2023)}]{srivastava2023big-bench}
BIG bench authors. 2023.
\newblock \href {https://openreview.net/forum?id=uyTL5Bvosj} {Beyond the
  imitation game: Quantifying and extrapolating the capabilities of language
  models}.
\newblock \emph{Transactions on Machine Learning Research}.

\bibitem[{Brown et~al.(2020)Brown, Mann, Ryder, Subbiah, Kaplan, Dhariwal,
  Neelakantan, Shyam, Sastry, Askell, Agarwal, Herbert-Voss, Krueger, Henighan,
  Child, Ramesh, Ziegler, Wu, Winter, Hesse, Chen, Sigler, Litwin, Gray, Chess,
  Clark, Berner, McCandlish, Radford, Sutskever, and Amodei}]{brown2020GPT3}
Tom~B. Brown, Benjamin Mann, Nick Ryder, Melanie Subbiah, Jared Kaplan,
  Prafulla Dhariwal, Arvind Neelakantan, Pranav Shyam, Girish Sastry, Amanda
  Askell, Sandhini Agarwal, Ariel Herbert-Voss, Gretchen Krueger, Tom Henighan,
  Rewon Child, Aditya Ramesh, Daniel~M. Ziegler, Jeffrey Wu, Clemens Winter,
  Christopher Hesse, Mark Chen, Eric Sigler, Mateusz Litwin, Scott Gray,
  Benjamin Chess, Jack Clark, Christopher Berner, Sam McCandlish, Alec Radford,
  Ilya Sutskever, and Dario Amodei. 2020.
\newblock \href {http://arxiv.org/abs/2005.14165} {Language models are few-shot
  learners}.

\bibitem[{Gunasekar et~al.(2023)Gunasekar, Zhang, Aneja, Mendes, Del~Giorno,
  Gopi, Javaheripi, Kauffmann, de~Rosa, Saarikivi
  et~al.}]{gunasekar2023textbooks}
Suriya Gunasekar, Yi~Zhang, Jyoti Aneja, Caio C{\'e}sar~Teodoro Mendes, Allie
  Del~Giorno, Sivakanth Gopi, Mojan Javaheripi, Piero Kauffmann, Gustavo
  de~Rosa, Olli Saarikivi, et~al. 2023.
\newblock Textbooks are all you need.
\newblock \emph{arXiv preprint arXiv:2306.11644}.

\bibitem[{Hendrycks et~al.(2020)Hendrycks, Burns, Basart, Zou, Mazeika, Song,
  and Steinhardt}]{hendrycks2020MMLU}
Dan Hendrycks, Collin Burns, Steven Basart, Andy Zou, Mantas Mazeika, Dawn
  Song, and Jacob Steinhardt. 2020.
\newblock Measuring massive multitask language understanding.
\newblock \emph{arXiv preprint arXiv:2009.03300}.

\bibitem[{Huang et~al.(2023)Huang, Bai, Zhu, Zhang, Zhang, Su, Liu, Lv, Zhang,
  Lei, Fu, Sun, and He}]{huang2023ceval}
Yuzhen Huang, Yuzhuo Bai, Zhihao Zhu, Junlei Zhang, Jinghan Zhang, Tangjun Su,
  Junteng Liu, Chuancheng Lv, Yikai Zhang, Jiayi Lei, Yao Fu, Maosong Sun, and
  Junxian He. 2023.
\newblock C-eval: A multi-level multi-discipline chinese evaluation suite for
  foundation models.
\newblock \emph{arXiv preprint arXiv:2305.08322}.

\bibitem[{Li et~al.(2023)Li, Hammoud, Itani, Khizbullin, and
  Ghanem}]{li2023camel}
Guohao Li, Hasan Abed Al~Kader Hammoud, Hani Itani, Dmitrii Khizbullin, and
  Bernard Ghanem. 2023.
\newblock \href {http://arxiv.org/abs/2303.17760} {Camel: Communicative agents
  for "mind" exploration of large scale language model society}.

\bibitem[{Lin and Chen(2023)}]{lin2023llmeval}
Yen-Ting Lin and Yun-Nung Chen. 2023.
\newblock \href {http://arxiv.org/abs/2305.13711} {Llm-eval: Unified
  multi-dimensional automatic evaluation for open-domain conversations with
  large language models}.

\bibitem[{Liu et~al.(2023{\natexlab{a}})Liu, Ning, Teng, Liu, Zhou, and
  Zhang}]{liu2023reasonAbility}
Hanmeng Liu, Ruoxi Ning, Zhiyang Teng, Jian Liu, Qiji Zhou, and Yue Zhang.
  2023{\natexlab{a}}.
\newblock \href {http://arxiv.org/abs/2304.03439} {Evaluating the logical
  reasoning ability of chatgpt and gpt-4}.

\bibitem[{Liu et~al.(2023{\natexlab{b}})Liu, Yang, Jia, Zhang, Zhou, Dai, Yang,
  and Vosoughi}]{liu2023stableAlign}
Ruibo Liu, Ruixin Yang, Chenyan Jia, Ge~Zhang, Denny Zhou, Andrew~M. Dai, Diyi
  Yang, and Soroush Vosoughi. 2023{\natexlab{b}}.
\newblock \href {http://arxiv.org/abs/2305.16960} {Training socially aligned
  language models in simulated human society}.

\bibitem[{Nair et~al.(2023)Nair, Schumacher, Tso, and Kannan}]{nair2023dera}
Varun Nair, Elliot Schumacher, Geoffrey Tso, and Anitha Kannan. 2023.
\newblock \href {http://arxiv.org/abs/2303.17071} {Dera: Enhancing large
  language model completions with dialog-enabled resolving agents}.

\bibitem[{Nijkamp et~al.(2023)Nijkamp, Pang, Hayashi, Tu, Wang, Zhou, Savarese,
  and Xiong}]{nijkamp2023codegen}
Erik Nijkamp, Bo~Pang, Hiroaki Hayashi, Lifu Tu, Huan Wang, Yingbo Zhou, Silvio
  Savarese, and Caiming Xiong. 2023.
\newblock \href {http://arxiv.org/abs/2203.13474} {Codegen: An open large
  language model for code with multi-turn program synthesis}.

\bibitem[{OpenAI(2023)}]{openai2023gpt4}
OpenAI. 2023.
\newblock \href {http://arxiv.org/abs/2303.08774} {Gpt-4 technical report}.

\bibitem[{Park et~al.(2023)Park, O'Brien, Cai, Morris, Liang, and
  Bernstein}]{park2023generative}
Joon~Sung Park, Joseph~C. O'Brien, Carrie~J. Cai, Meredith~Ringel Morris, Percy
  Liang, and Michael~S. Bernstein. 2023.
\newblock \href {http://arxiv.org/abs/2304.03442} {Generative agents:
  Interactive simulacra of human behavior}.

\bibitem[{Park et~al.(2022)Park, Popowski, Cai, Morris, Liang, and
  Bernstein}]{park2022socialSim}
Joon~Sung Park, Lindsay Popowski, Carrie~J. Cai, Meredith~Ringel Morris, Percy
  Liang, and Michael~S. Bernstein. 2022.
\newblock \href {http://arxiv.org/abs/2208.04024} {Social simulacra: Creating
  populated prototypes for social computing systems}.

\bibitem[{Premack and Woodruff(1978)}]{premack1978ToM}
David Premack and Guy Woodruff. 1978.
\newblock Does the chimpanzee have a theory of mind?
\newblock \emph{Behavioral and brain sciences}, 1(4):515--526.

\bibitem[{Qian et~al.(2023)Qian, Cong, Yang, Chen, Su, Xu, Liu, and
  Sun}]{qian2023chatdev}
Chen Qian, Xin Cong, Cheng Yang, Weize Chen, Yusheng Su, Juyuan Xu, Zhiyuan
  Liu, and Maosong Sun. 2023.
\newblock \href {http://arxiv.org/abs/2307.07924} {Communicative agents for
  software development}.

\bibitem[{Shen et~al.(2023)Shen, Song, Tan, Li, Lu, and
  Zhuang}]{shen2023hugginggpt}
Yongliang Shen, Kaitao Song, Xu~Tan, Dongsheng Li, Weiming Lu, and Yueting
  Zhuang. 2023.
\newblock Hugginggpt: Solving ai tasks with chatgpt and its friends in
  huggingface.
\newblock \emph{arXiv preprint arXiv:2303.17580}.

\bibitem[{Sun et~al.(2023)Sun, Zhang, Deng, Cheng, and
  Huang}]{sun2023safetyAssess}
Hao Sun, Zhexin Zhang, Jiawen Deng, Jiale Cheng, and Minlie Huang. 2023.
\newblock \href {http://arxiv.org/abs/2304.10436} {Safety assessment of chinese
  large language models}.

\bibitem[{Wang et~al.(2023{\natexlab{a}})Wang, Xie, Jiang, Mandlekar, Xiao,
  Zhu, Fan, and Anandkumar}]{wang2023voyager}
Guanzhi Wang, Yuqi Xie, Yunfan Jiang, Ajay Mandlekar, Chaowei Xiao, Yuke Zhu,
  Linxi Fan, and Anima Anandkumar. 2023{\natexlab{a}}.
\newblock \href {http://arxiv.org/abs/2305.16291} {Voyager: An open-ended
  embodied agent with large language models}.

\bibitem[{Wang et~al.(2023{\natexlab{b}})Wang, Liang, Meng, Sun, Shi, Li, Xu,
  Qu, and Zhou}]{wang2023chatgptNLGeval}
Jiaan Wang, Yunlong Liang, Fandong Meng, Zengkui Sun, Haoxiang Shi, Zhixu Li,
  Jinan Xu, Jianfeng Qu, and Jie Zhou. 2023{\natexlab{b}}.
\newblock \href {http://arxiv.org/abs/2303.04048} {Is chatgpt a good nlg
  evaluator? a preliminary study}.

\bibitem[{Weng(2023)}]{weng2023AutoAgents}
Lilian Weng. 2023.
\newblock \href {https://lilianweng.github.io/posts/2023-06-23-agent/}
  {Llm-powered autonomous agents}.
\newblock \emph{lilianweng.github.io}.

\bibitem[{Yao et~al.(2023)Yao, Yu, Zhao, Shafran, Griffiths, Cao, and
  Narasimhan}]{yao2023ToT}
Shunyu Yao, Dian Yu, Jeffrey Zhao, Izhak Shafran, Thomas~L. Griffiths, Yuan
  Cao, and Karthik Narasimhan. 2023.
\newblock \href {http://arxiv.org/abs/2305.10601} {Tree of thoughts: Deliberate
  problem solving with large language models}.

\bibitem[{Zhong et~al.(2023)Zhong, Cui, Guo, Liang, Lu, Wang, Saied, Chen, and
  Duan}]{zhong2023agieval}
Wanjun Zhong, Ruixiang Cui, Yiduo Guo, Yaobo Liang, Shuai Lu, Yanlin Wang, Amin
  Saied, Weizhu Chen, and Nan Duan. 2023.
\newblock \href {http://arxiv.org/abs/2304.06364} {Agieval: A human-centric
  benchmark for evaluating foundation models}.

\bibitem[{Zhou et~al.(2023)Zhou, Liu, Xu, Iyer, Sun, Mao, Ma, Efrat, Yu, Yu,
  Zhang, Ghosh, Lewis, Zettlemoyer, and Levy}]{zhou2023lima}
Chunting Zhou, Pengfei Liu, Puxin Xu, Srini Iyer, Jiao Sun, Yuning Mao, Xuezhe
  Ma, Avia Efrat, Ping Yu, Lili Yu, Susan Zhang, Gargi Ghosh, Mike Lewis, Luke
  Zettlemoyer, and Omer Levy. 2023.
\newblock \href {http://arxiv.org/abs/2305.11206} {Lima: Less is more for
  alignment}.

\end{thebibliography}
\bibliographystyle{acl_natbib}




\end{document}